# Deep Reinforcement Learning Based Systems for Safety Critical Applications in Aerospace

Abedin Sherifi
*December 2024*

*Abstract*—Recent advancements in artificial intelligence (AI) applications within aerospace have demonstrated substantial growth, particularly in the context of control systems. As High Performance Computing (HPC) platforms continue to evolve, they are expected to replace current flight control or engine control computers, enabling increased computational capabilities. This shift will allow real-time AI applications, such as image processing and defect detection, to be seamlessly integrated into monitoring systems, providing real-time awareness and enhanced fault detection and accommodation.

Furthermore, AI's potential in aerospace extends to control systems, where its application can range from full autonomy to enhancing human control through assistive features. AI, particularly deep reinforcement learning (DRL), can offer significant improvements in control systems, whether for autonomous operation or as an augmentative tool.

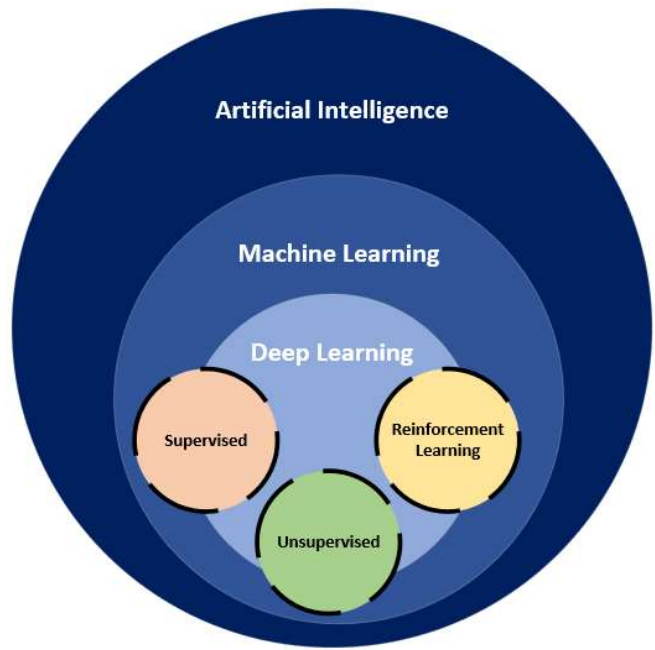

Fig. 1: AI Description

## I. Introduction

AI is a vast field encompassing numerous subdomains, all aimed at replicating human cognitive functions. Machine Learning (ML), a key subset of AI, involves algorithms that improve their performance as they process more data. Deep Learning (DL), a further specialization of ML, employs multi-layer neural networks to analyze large datasets. A diagram that captures the overall AI field is presented below in Figure 1.

In AI, Supervised Learning utilizes labeled datasets for regression and classification tasks, while Unsupervised Learning focuses on unlabeled datasets for clustering. This paper, however, centers on DRL, which is particularly relevant for control and decision-making tasks in dynamic environments.

The main distinction between machine learning algorithms and classical control methods lies in the stochastic nature of ML models, as opposed to the deterministic nature of traditional algorithms. ML systems derive their behavior from data, while classical control systems follow predefined instructions.

Reinforcement Learning (RL) involves learning by interacting with an environment, rather than using a static dataset. The primary goal of RL is to discover a set of actions that maximize long-term outcomes. DRL, which combines RL with deep learning, has been successfully applied in games such as Atari and Go, as well as in applications like Tesla's motion planning for path-finding and obstacle avoidance [1], where deep neural networks approximate Q-values to determine optimal actions.

A. Sherifi is with the Artificial Intelligence Department, University of Texas at Austin. A. Sherifi is with Pratt and Whitney, a RTX Company

## II. AI-BASED SYSTEM DEVELOPMENT GUIDANCE VIA CURRENT STANDARDS

### A. Overall Summary

The use of AI-based systems in safety critical applications is very hard to achieve through existing regulatory standards for Software and Hardware. Existing standards such as ARP-4754 [2], DO-178C [3], DO-254 [4] fail to provide compliance for AI based systems.

The purpose of aircraft certification is to minimize and prevent accidents. Each set of standards is designed to reduce the likelihood of aircraft failures. The established procedures for traditional aerospace system safety and certification level determination are outlined in ARP-4754 and ARP-4761 [2] [5]. These processes guide manufacturers in determining the necessary certification level for both hardware and software. DO-254 is the widely accepted certification standard for hardware, while DO-178C serves the same purpose for software. DO-178C emphasizes the principle of ensuring that every line of code is traced to a requirement, and vice versa, guaranteeing that each line of code serves a specific purpose and that all requirements are met [3] [4].

One approach to verifying AI-based systems, specifically DRL systems, is the following from [6]:
- **Verification during Exploration** – Ensuring that the RL system satisfies predefined properties P within an environmental model M while exploring the environment E using a policy P.
- **Verification during Exploitation** – Validating that the RL system adheres to properties P when exploiting the policy P in the environment E.
- **Validation of Verification Model** – Confirming that the model M accurately represents the environment E with an acceptable level of accuracy A.
- **Validation of Performance** – Verifying that the total reward R obtained from exploiting policy P in the target environment, $T_E$, when starting in state, S, is within an error margin, D, of the total reward, R, when exploiting target policy, $T_P$, in training environment, E, when starting in state, S.

The steps above are key to tackling the issue of formally being able to validate and verify DRL requirements. The steps above break up the whole DRL algorithm into specific parts that can be individually validated and/or verified.

One approach to utilizing AI based systems for safety critical applications would be to complement such systems with emergency back-up procedures that can be implemented via the use of existing standards such as ARP-4754, DO-178, and/or DO-254. This is further explained in 6.

### B. Current Standards and On-Going Initiatives by EASA and FAA

A few years ago, the European Union Aviation Safety Agency (EASA), in collaboration with SAE International, began working on the implementation of a new standard for AI applications in the aviation industry. EASA's AI roadmap is specifically focused on Level 1 AI, which is designed to assist humans. The development of ARP6983 (Process Standard for Development and Certification/Approval of Aeronautical Safety-Related Products Implementing AI) is currently underway, with guidelines being established specifically for Level 1 AI applications.

To drive the certification of AI-based systems, EASA introduced several High-Level Properties (HLPs). The main HLPs are listed below [7] [8]:

- **Auditability**: The degree to which an independent examination of the development and verification process can be conducted.
- **Data Quality**: The extent to which data are accurate, free from defects, and possess the desired features.
- **Explainability**: The extent to which the behavior of a Machine Learning model can be understood by humans.
- **Maintainability**: The ability to extend or improve the system while preserving compliance with unchanged requirements.
- **Resilience**: The system's ability to continue functioning in the presence of an error or fault.
- **Robustness**: (Global) The ability of the system to perform its intended function despite abnormal or unknown inputs; (Local) the consistency of responses for similar inputs.
- **Specifiability**: The extent to which the system can be fully and accurately described through a list of requirements.
- **Verifiability**: The ability to evaluate whether the system meets its requirements (adapted from ARP4754A).

Figure 2 below illustrates one potential framework for certifying AI-based systems, based on Engineering Decisions, Engineering Practices, Machine Learning HLPs, and overall HLPs.

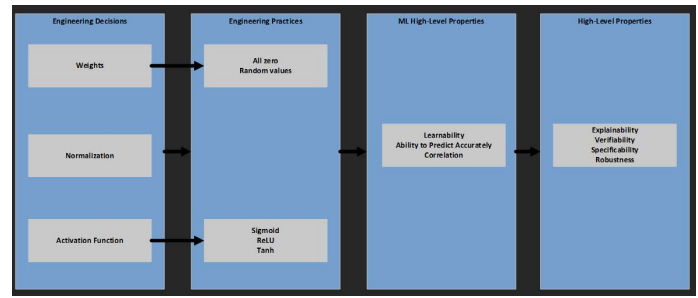

Fig. 2: EASA HLPs [7] [8]

The table I below summarizes the EASA AI Roadmap AI Levels for reference [7].

| EASA AI Roadmap AI Levels | Function allocated to the system to contribute to the high-level task |
|---|---|
| Level 1A Human Augmentation | Automation support to information acquisition and analysis |
| Level 1B Human Assistance | Automation support to decision-making |
| Level 2 Human-AI Collaboration | Overseen and overridable automatic decision-making and action implementation |
| Level 3A More Autonomous AI | Overridable automatic decision-making and action implementation |
| Level 3B Fully Autonomous AI | Non-overridable automatic decision-making and action implementation |

TABLE I: EASA AI Levels [7]

Figure 3 below presents the amended V-Model for safety-critical systems with AI, based on EASA's AI Roadmap research [7]. During the training phase, iterations are carried out for requirements related to AI/ML, data management, learning process management, model training, and learning process verification. In the implementation phase, the following steps are carried out: model implementation, inference model verification and integration, independent data and learning verification, and AI/ML constituent requirements verification. This amended V-Model is significantly different from the traditional software V-Model.

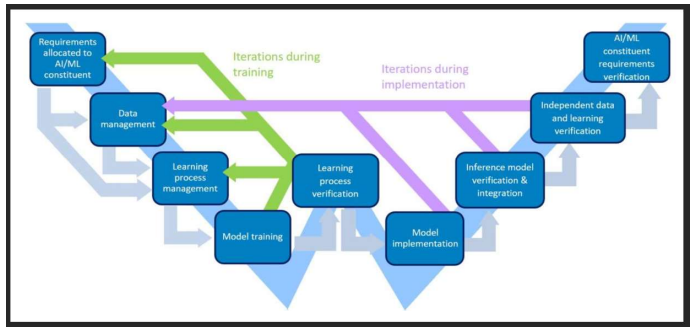

Fig. 3: EASA's W-Model for AI-Based Systems [7]

NASA with several aerospace companies has also worked on a framework that can be applied as a guide to ML functions. That framework is called the Overarching Properties (OPs). The OPs consist of 3 key attributes: intent, correctness, and innocuity. In Europe, EASA has also worked on similar path and they have called their framework the Abstraction Layers. Both, the OPs and Abrstraction Layers are based on methods of good argumentation to certify AI based systems.

Based on the work that EASA, FAA, NASA, and rest of the aerospace community have done, it is possible to deploy AI models for safety critical systems as long as requirements are specifically captured, argumented with evidence, and used in parallel with backup procedures.

DO-333 is a supplement to DO-178C that provides guidance on ensuring airworthiness for components used in aerospace applications. Figure 4 below illustrates the typical software requirements process followed for such components. It begins with system requirements at the top, which are then broken down into high-level requirements (HLRs). These HLRs are further decomposed into low-level requirements (LLRs) and software architecture. From there, source code and executable binary files are generated. Testing based on these requirements is performed against both the HLRs and LLRs.

One challenge in the context of ML models is structural coverage, which is a critical aspect of testing. Structural coverage helps identify unintended functionality, dead code, or missing requirements in traditional software. However, applying this concept to ML models is problematic due to the inherent complexity and adaptability of these models. In ML, structural coverage may not effectively capture all aspects of the model's behavior, making it more difficult to ensure complete testing and validation.

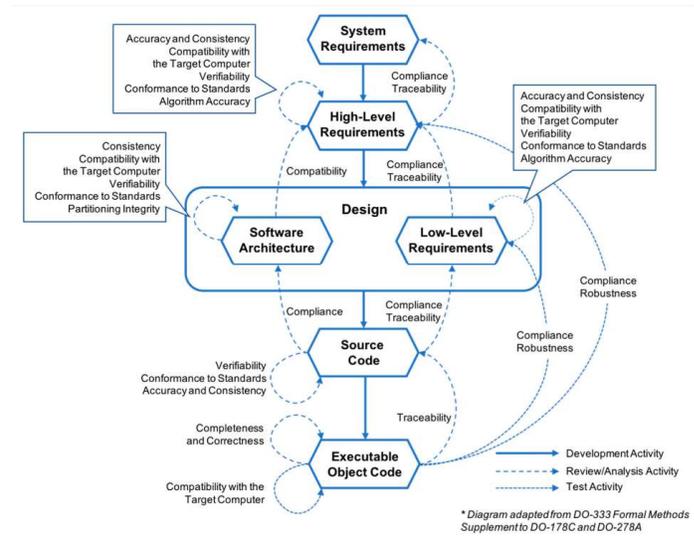

Fig. 4: DO-333 Design Assurance Process [9]

Figure 5 below illustrates one of the main challenges in certifying ML models: capturing the traceability from the ML model architecture and datasets to the trained model with specific weights is a difficult task. The difficulty in establishing traceability between the ML model architecture, dataset, and trained model with weights arises from the complexity, lack of transparency, randomness, and large scale of modern machine learning systems.

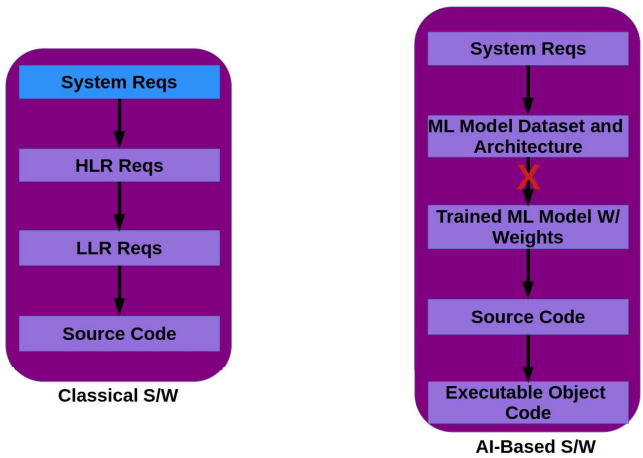

Fig. 5: Requirement Traceability for Classical vs. AI-Based Systems

Challenges specific for ML certification are the following:
- Requirements-based testing necessitates clearly defined requirements, which are difficult to capture for every element of an ML model.
- Formal verification of requirements also requires well-defined requirements, which are often not available.
- ML model components such as layers, neurons, weights, and individual lines of code cannot be directly traced to specific requirements.
- Structural coverage is not easily applicable to ML models.
- Demonstrating the absence of unintended behavior in an ML model is challenging.

### III. Deep Reinforcement Learning AI-Based Systems in Aerospace

The RL algorithm is based on the Bellman equation. The Bellman equation states that long term reward per an action is equal to the reward from the current action and expected reward for future actions. The Bellman equation is given below:

$$V(s) = max_a R(s,a) + \gamma V(s')  \quad (1)$$

The value of a given state is equal to the max action which implies that the state with maximized values is chosen. Next, the V value for the next state is taken multiplied by a discount factor which dictates the importance of the reward over time and adds it to the value function V.

RL algorithms can be generally applied to uncertain dynamics since they do not rely on beforehand on a dynamic model. Typically, any RL agent can either act greedy with the available data which is part of exploitation, or explore which ultimately means the agent taking suboptimal actions to learn a more accurate model which is called exploration.

The fundamental piece of RL is the RL agent learning through interaction with the environment. The RL agent observes the result of its actions based on the environment feedback and then decides to alter its future actions in order to maximize cumulative rewards. The RL agent observes a state called $s_t$ at time $t$. The RL agent interacts with the environment by taking action $a_t$ at time $t$ in state $s_t$. When the RL agent takes this action, the agent and the environment transition to a new state called $s_{t+1}$. The best steps of actions for the RL agent to take are governed by the rewards provided by the environment. The goal of the RL agent is to learn a policy or a control schema $\pi$ that can maximize the cumulative rewards. [10]

In mathematical terms, the RL algorithm can best be described as a Markov Decisions Process (MDP) [10]:
- A set of states **S**.
- A set of actions **A**.
- The mapping of state-action pair at time $t$ onto a distribution of states at time $t+1$.
- A reward function $R(s_t, a_t, s_{t+1})$.
- A discount factor $\gamma \in [0,1]$ with lower values placing more emphasis on immediate rewards and higher values placing more emphasis on long-term rewards.

There are two distinct methods of training a DRL agent: offline and online learners. An off-line DRL agent is trained offline in the training environment, and then it will use the policy with the maximum reward in the deployment or target environment based on the training environment. This implies that the agent will not change its policy in the deployment environment. For the on-line agent, on the other hand, this implies that the agent will also get trained in the deployment environment. For such on-line training agents, it is wise to verify the agent in run-time [10].

One of the critical drawbacks of online training agent is that the agent will perform trial and errors on the deployment target which could jeopardize the life of those on-board the aircraft. There is a trade-off between deciding to go with an online trained agent or an offline trained agent. The offline trained agent will not be able to train in the deployment environment and optimize furthermore its policy. On the other hand, the online trained agent will be able to train in its deployed environment, but it could also jeopardize the safety of those on board since part of this online training is exploration which could imply random un-safe actions from the agent [10].

One of the key drawbacks of training the agent in a simulated environment is dealing with model inaccuracies. However, training in a simulated environment is great for speed, safety, and flexible conditions. The drawback for training an agent in a real environment is risk, but accuracy is higher that a simulated model training.

RL algorithms in the past have suffered from memory complexity and computational complexity. But, advances in DL have allowed for RL applications to rely on DNN function approximators.

It is of great importance to mention that the actions taken by the RL agent are sequential in nature and depend on the previous actions and states.

For AI-based systems, runtime assurance switches play a crucial role in detecting and mitigating unsafe or unexpected behaviors during operation. These runtime monitors continuously observe the inputs, outputs, and internal states of an ML model, enabling the system to switch to a safer behavior or activate a recovery function when an issue is detected. This mechanism ensures that, even if the ML model itself could lead to unintended behaviors due to its complexity or limitations, the overall system can still maintain safety by transitioning to a safer state with minimal performance degradation [11].

The use of runtime assurance monitors allows for the deployment of ML models in safety-critical systems, addressing the risks associated with unpredictable model behaviors. By incorporating safety backup components and a switch mechanism between the primary ML model and these backup systems, it is possible to maintain safety while benefiting from the capabilities of advanced ML models. Runtime monitors can include various components like captured below:

- **System Safety Monitors**: To ensure that the overall system remains in a safe operational state.
- **Deep Neural Network Confidence Assessment**: To evaluate the confidence level of the model's predictions and trigger intervention when confidence falls below a threshold.
- **Monitor Selectors**: These can assess which monitoring function is most appropriate for the current system state and behavior.
- **Contingency Managers**: These components help determine when to intervene or switch to backup systems, ensuring continued safe operation in case of detection of unsafe behavior.

Together, these runtime assurance mechanisms provide a layered approach to safety, making it feasible to deploy ML models in environments where safety is paramount.

The following diagram captures a Software partitioned-based approach to the implementation of DRL system within an aerospace application.

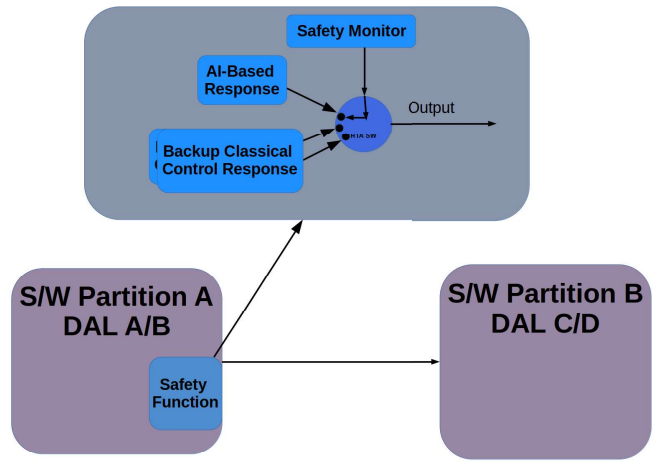

Fig. 6: Run-Time Assurance Switch

The following W-Model is something that we could use for the qualification and certification of DRL-based systems.

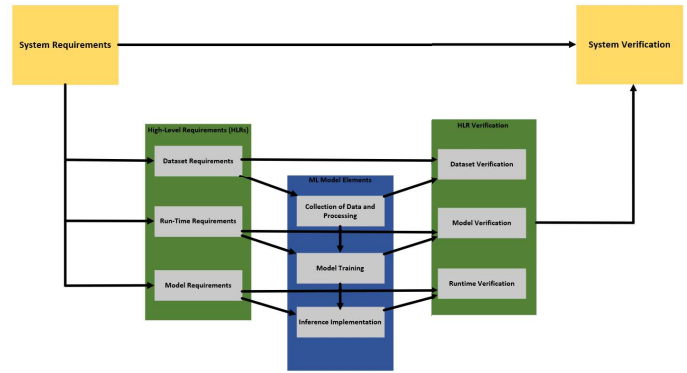

Fig. 7: W-Model for DRL-Based System

The report referenced in [12] covers in detail the use of OPs to provide sufficient suitability of a product for installation on an aircraft. If we dive further into the OPs, the three main properties of OPs are the following:

- Intent: The defined intended behavior is correct and complete with respect to the desired behavior.
- Correctness: The implementation is correct with respect to its defined intended behavior, under foreseeable operating conditions.
- Innocuity: Any part of the implementation that is not required by the defined intended behavior has no acceptable impact.

Innocuity here is very similar to the SOTIF area in automotive industry that covers the impact of unintended functionality.

For the case of qualifying and certifying a DRL application for controls, the three figures below capture the overall process involved if OPs are followed for AI specific components and existing approach (ARP-4754, DO-254, DO-333, DO-178) is used for non-AI components.

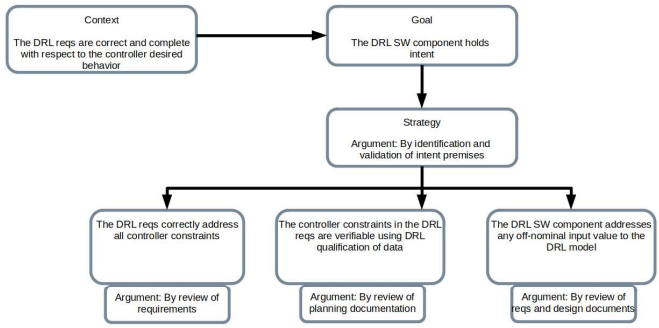

Fig. 8: Assurance Details for the DRL SW Component Holds Intent

Figure 8 covers the process of showing that the DRL control system behavior is correct and complete with respect to the desired behavior.

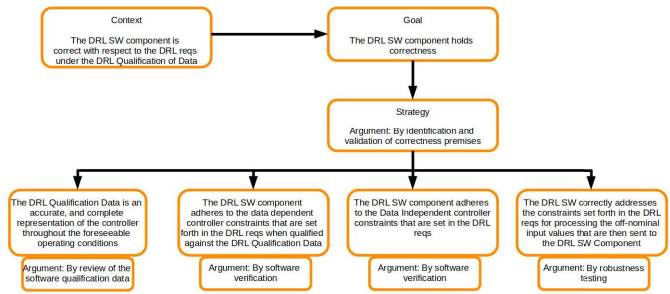

Fig. 9: Assurance Details for the DRL SW Component Holds Correctness

Figure 9 covers the process of showing that the DRL control system implementation is correct with respect to the defined intended behavior under foreseeable operating conditions.

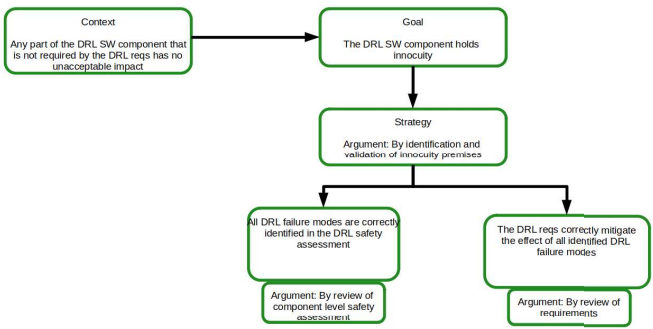

Fig. 10: Assurance Details for the DRL SW Component Holds Innocuity

Figure 10 covers the process of showing that any part of the DRL control system implementation that is not required by the defined intended behavior has no unacceptable impact.

## IV. Classical vs. DRL-Based Controllers

The table II below captures the comparison of the key parts making up the DRL controller and a classical controller. This is shown to aid the reader for better understanding the key parts enabling the use of DRL.

| DRL | Classical Control |
|---|---|
| *Policy* | *Controller* |
| *Environment* | *Plant* |
| *Observation* | *Measurement* |
| *Action* | *ManipulatedVariable* |
| *Reward* | *Error/CostFunction* |

TABLE II: DRL and Classical Control Parts Naming

Figure 11 below illustrates the classical control representation, while Figure 12 below shows the DRL control representation. These figures are designed to help the user clearly distinguish the different components that constitute classical control and DRL.

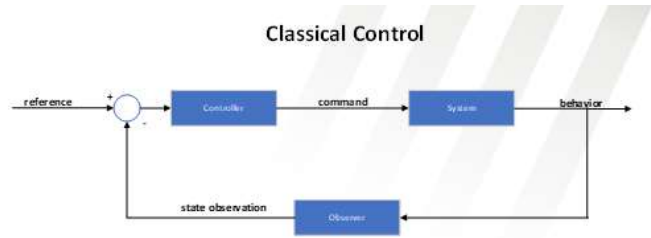

Fig. 11: Classical Control

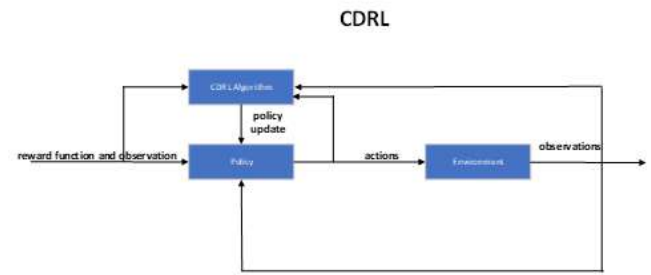

Fig. 12: DRL Control

Figure 13 below illustrates the key distinction between classical software control and AI-based Deep Reinforcement Learning control. In classical software control, the system operates by receiving a state as input and generating corresponding actions as output based on predefined logic or rules. This process follows a fixed and deterministic approach, where the software directly maps states to actions.

In contrast, Deep Reinforcement Learning control involves a more dynamic process. Here, the learning agent engages in exploration to interact with the environment and gather data. This exploration is fed into the agent's policy, which also takes the current state as input. The policy then determines the appropriate action to take. The key difference is that in DRL, the agent continuously learns and refines its policy over time through experience, rather than relying on a static mapping between states and actions. This learning-based approach enables the system to adapt and improve its performance based on feedback from the environment, allowing for more complex and flexible decision-making processes.

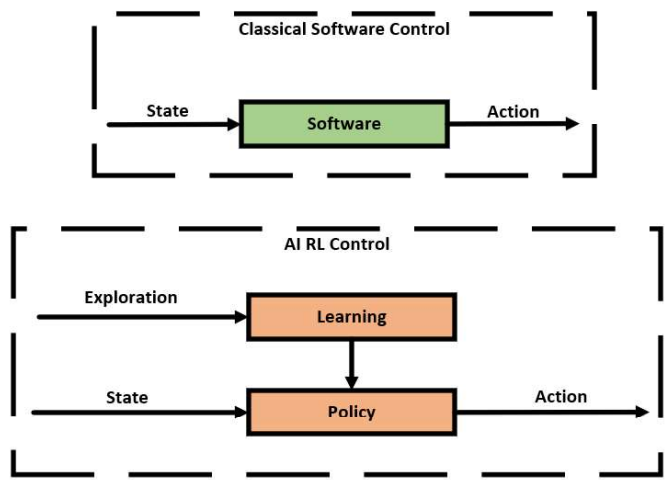

Fig. 13: Classical vs DRL Software Control